\documentclass{amia}
\usepackage{graphicx}
\usepackage{bbm}
\usepackage{amsmath,bm}
\usepackage[labelfont=bf]{caption}
\usepackage[style=numeric,sorting=none,sortcites=true,maxnames=6,minnames=3,autocite=superscript]{biblatex}
\usepackage{color}
\usepackage{tabularx}
\usepackage{booktabs}
\usepackage{longtable}
\usepackage{makecell}
\usepackage{multirow}
\usepackage{placeins}
\usepackage{setspace}

\begin{document}

\title{Learning Hierarchical Representations of Electronic Health Records\\ for Clinical Outcome Prediction}

\author{Luchen Liu$^{1}$,~~ Haoran Li$^{1}$,~~ Zhiting Hu$^{2}$,~~ Haoran Shi$^{1}$,\\ Zichang Wang$^{1}$,~~ 
Jian Tang, PhD$^{3,4}$,~~ Ming Zhang\footnote{Corresponding author.}, PhD$^{1}$}

\institutes{
    $^1$ Department of Computer Science, Peking University, Beijing, China \\
    $^2$Carnegie Mellon University, Pittsburgh, PA, US \\
     $^3$Mila - Qu\'ebec AI Institute, Montr\'eal,  Qu\'ebec, Canada \\
     $^4$HEC Montr\'eal,  Montr\'eal, Qu\'ebec, Canada \\
}

\maketitle

\noindent{\bf Abstract}

{\it
Clinical outcome prediction 
based on Electronic Health Record (EHR) helps enable early interventions for high-risk patients, and is thus a central task for smart healthcare. Conventional deep sequential models fail to capture the rich temporal patterns encoded in the long and irregular clinical event sequences in EHR.
%clinical outcome prediction is challenging due to 
% Objective:
%We aim to learn hierarchical representations of heterogeneous event sequences in EHR data to improve clinical outcome prediction.  
% Materials and Methods: 
%utilizing rich temporal information encoded in the long irregular clinical event sequences is hard via classic deep sequential models. 
%As the short-range orders of clinical events  can be ignored in EHR data,
%long-term dependencies in the original sequence can be captured through temporal dependencies of selected important events in each adaptively segmented group.
We make the observation that 
clinical events at a long time scale exhibit strong temporal patterns, while events within a short time period tend to be disordered co-occurrence. We thus propose differentiated  mechanisms to model clinical events at different time scales.
%Based on the observation that clinical events in different time scales 
%events occurring within a short period do not have a definite order, we can select critical events in short-range event groups as the basis for capturing long-term dependencies in the long time scale.
Our model learns hierarchical representations of event sequences, to adaptively distinguish between short-range and long-range events, and accurately capture their core temporal dependencies.
%using a recurrent neural network with both group-specific event attention and inter-group temporal attention.
% Results: 
Experimental results on real clinical data show that our model 
greatly improves over previous state-of-the-art models, achieving AUC scores of 0.94 and 0.90 for predicting death and ICU admission, respectively.
Our model also successfully identifies   important events for different clinical outcome prediction tasks. 
% Conclusion: 
%The hierarchical representations of heterogeneous event sequences achieve significantly improved performance on real clinical datasets for death and ICU admission prediction tasks, and provide deep insights for intelligent healthcare decisions.
}

\section{Introduction}
%\subsection{BACKGROUND AND SIGNIFICANCE}
The ever-growing massive Electronic Health Records (EHR) data expose an opportunity for large-scale data-driven health analysis and intelligent medical care. 
Predicting clinical outcomes, such as death and intensive care unit (ICU) admission, plays an important role in improving the performance of healthcare systems.
For instance, accurate clinical prediction based on patients‘ existing medical records can enable advanced and timely medical intervention.
%even if doctors do not recognize before the occurrence of serious outcomes.

%\subsection{temporal dependncy of the EHR data}
Clinical outcome prediction is challenging because it is hard to utilize rich temporal information encoded in the sequence of clinical events in EHR data~\autocite{choi2016using}. 
In particular, EHR data usually consist of clinical events with irregular intervals~\autocite{Choi2016:Multi} from heterogeneous sources, including patient health features (vital sign measurements, laboratory test results, etc), medical interventions (procedures, drug inputs, etc), and expert judgments (diagnoses, notes, etc)~\autocite{johnson2016mimic}. 
The temporal order of these events is critical for  predicting clinical outcome. 
For example, patient health features can be affected by previous medical interventions, and in turn determine subsequent medical interventions through expert judgments.  

%\subsection{relatedwork}
Conventional approaches that directly apply classic deep sequential models, such as recurrent neural networks~\autocite{pham2016deepcare,che2016recurrent} (RNN) and convolutional neural networks~\autocite{nguyen2016deepr} (CNN), usually fail to capture temporal dependencies in such long irregular event sequences~\autocite{Karlsson2016Predicting, lipton2016directly},
%is hard, via directly applied deep sequential models. 
%However, these models did not capture the multi-scale temporal dependencies of EHR data.
%because events in the long sequences are modeled one by one.  
as long-term dependencies can easily exceed the modeling capacity. 
%of recurrent neuron networks (RNN) or convolution networks (CNN).
To handle irregularly timed events, some extensions of the classic models have been developed, such as time-aware RNN~\autocite{baytas2017patient} or CNN~\autocite{nguyen2016mathtt}.
However, their performance is still largely unsatisfying due to the limited ability to capture long-term dependencies.       
%Some work used convolution Nonnegative Matrix Factorization(cNMF) models~\autocite{Wang2013:A}.
%But these methods pay much attention to short-range order of the event sequence, because the useless short-range order of events is emphasized by the direct application of recursive units and convolution structures on fine grained clinical events. 

%These medical features can be organized as many kinds of clinical events with heterogeneous attributes. For example, an event can consist of one categorical attribute, i.e. the type of medical solution, and two numerical attributes, i.e., input rate and medicine dose. Clinical events are heterogeneous with each other, because different types of events may have different attributes, and 

%To maintain the temporal information in the EHR data of a patient can be represented as a sequence of heterogeneous events for predictive tasks~\cite{}

%Furthermore, different events of the same type can also be different because of the various value of numerical  and categorical attributes of the events. Therefore,  it is desirable to design an expressive yet clean representation for EHR data in order to retain the heterogeneous information of clinical events.

%\subsection{ignord short-range order  of EHR data}
%The challenging characteristic of EHR data is multi-scale temporal dependency
%of heterogeneous event sequences.
This work aims to address the above challenges. We first make a key observation that, though the clinical events in EHR data can exhibit strong temporal patterns at a \emph{long} time scale, the events occurring within a \emph{short} time period usually do not have a definite order.
%temporally unordered.
%in EHR data, the short-range orders of clinical events can be ignored. 
%The timestamps of clinical events are irregular, and the short-term dependencies of events at small scales is not the same as long-term dependencies at large scales.
Specifically, unlike word sequences in natural languages where word tokens are ordered by grammar rules, clinical events recorded in a short period of time are instead a series of events, such as clinical laboratory test results,  that
%typically do not have definite order. 
%In fact, a series of clinical laboratory test results, recorded in a short period of time, 
only reflect the patient’s status in different views. 
%many clinical events are recorded with the same timestamp, and
Therefore, direct temporal modeling of such short-range events as in previous work can introduce noise and harm the temporal predictive performance. 
%there is not much predictive information in the short-range order of EHR data. 
Instead, local dependencies of these events should be modeled as event co-occurrence, and we can further select critical events from each of these short-range event groups as the basis for modeling the real temporal dependencies at a long time scale.
%and model the temporal dependencies of these event groups in a shorter sequence.
A key difficulty to this end is that the criterion for distinguishing long-term temporal dependencies from the local co-occurrence of critical events in a short range can vary across different diseases and phases, especially in the irregular EHR data~\autocite{ghassemi2015multivariate, leebig}. 

%retain shortage
%
%It is, therefore, desirable to adaptively treat the heterogeneous clinical event sequences in a hierarchical manner for outcome prediction tasks, so that the local dependency without short-range order of events can be modeled in low-level represents of the model, meanwhile the long-term dependency can be captured in the high-level representations.

To address the difficulties mentioned above, 
we propose a hierarchical neural network for clinical outcome prediction. 
Specifically, 
%we embed the clinical events with heterogeneous attributes in a low-dimensional space. 
%to learn hierarchical representations,
we adaptively segment irregular event sequences into sequential groups to distinguish short-range co-occurrence and long-term temporal dependencies as well as to learn hierarchical representations of the event sequence. 
At the low level, the model automatically identifies critical clinical events in each group and aggregates the events to form event group representations. 
At the high level, meaningful long-term dependencies of clinical event groups are captured in a sequence representation vector by a recurrent neural network.
Compared to traditional methods, the proposed method has several advantages:
\begin{itemize}
    \item  Our model can deal with the temporal irregularity of clinical event sequences by adaptively segmenting an event sequence into sequential groups.
    \item Our model learns a hierarchical representation of long and irregular event sequences to capture long-term dependencies of clinical events.
    \item The model is capable of discovering critical event groups as well as critical events in each group, through a temporal attention and event attention mechanism. This provides useful clinical insights for accurate prediction.
\end{itemize}

\section{Related Works}
\subsection{Modeling EHR Data}
Most existing works based on EHR data have either focused on stationary clinical text~\autocite{gao2017hierarchical,shi2017towards} and images~\autocite{greenspan2016guest,li2018hybrid,li2019knowledge}, or ignored irregular time intervals of temporal clinical events~\autocite{Liu2015Temporal,Henriksson2015Modeling,huddar2016predicting}. 
For example, previous work trained the semantic embeddings for the categories of clinical events for adverse drug event detection~\autocite{Henriksson2015Modeling}, or proposed a multi-view learning method that generalizes Canonical Correlation Analysis for an arbitrary collection of matrices involving missing data~\autocite{huddar2016predicting}. 
These works make predictions based on the   clinical events with regular time intervals, and  cannot distinguish short-range order from long-term temporal order of different diseases and patients. 
Our work addresses the issue by adaptive segmentation of clinical event sequences.

As long-term temporal dependencies are hard to capture, many works use a small subset of the whole EHR information, to avoid dealing with the long clinical event sequences. 
Some works select a subset of the numerical clinical features (the numerical attributes of clinical events) in the EHR data according to the expertise of clinicians~\autocite{che2016recurrent}. 
For instance,  Yoon
only uses a set of 21 (temporal) physiological streams comprising a set of 11 vital signs and 10 lab test scores to predict ICU admission~\autocite{yoon2016forecasticu}. Some works used graphical models to model patients’ health status~\autocite{caballero2015dynamically}. 
Some techniques transform selected 99 time series features of all the EHR data into a new latent space using the hyper-parameters of multi-task GP (MTGP) models to model patient  similarity~\autocite{choi2016using}. 
Recently, RETAIN used two reversed recursive neural networks (RNN) generating attention variables of sequential international disease classification (ICD) code groups for the prediction~\autocite{choi2016retain}. However, the codes are grouped by the fixed-length time slots for distinct patients and diseases, and local dependencies and long-term dependencies may be mixed up.
But these works can lose significant information, due to the expert bias when selecting a limit fraction of all clinical features in EHR as the input of the models, and fail to provide new data-driven insights for better healthcare.

\subsection{Clinical Outcome Prediction}
The clinical outcome prediction problem is studied by many works. However  many of them cannot take advantage of the temporal information in EHR data for prediction. Some of these studies used latent variable models to decompose bag-of-words free-text extracted from clinical event descriptions into meaningful features to predict patient mortality~\autocite{ghassemi2014unfolding}. ``Deep patient''~\autocite{miotto2016deep} arranged all clinical descriptors (features) in a period of time in a sparse vector without temporal information and trained the deep representation of patients with a 3 layer denoising autoencoder for diagnosis. Some work studied how to diagnose and predict Alzheimer’s disease (AD) with a hybrid manifold learning for non-temporal clinical feature embedding and the bootstrap aggregating (Bagging) algorithm~\autocite{dai2016bagging}. There is also a work model EHR data by factorizing the medical feature matrix into a latent medical concept mapping matrix and a concept value evolution matrix, and then they averaged all vectors in the evolution matrix to predict heart failure~\autocite{Zhou2014From}. 
Our model learns the hierarchical representations of clinical event sequences to utilize the temporal information for clinical outcome prediction.
\section{Data and Task Descriptions}
We give the notations and data descriptions of the predictive tasks in the following.

\subsection*{Clinical Events in EHR}
A clinical event is a record in the database of EHR, which describes a clinical activity of a particular patient at a certain time. The events can be  measurements of vital signals, injection of drugs, results of laboratory tests, and so on, which are summarized in table \ref{tab:clinical events}.
Each clinical event has some attributes, including categorical attributes and numerical attributes.
 For example, the lab test event $e_t$ has 2 categorical features and 1 numerical feature: $e_t=[Lab Item: Cholesterol, Abnormal Label: Abnormal, Result Index: 51\mu  /L]$. The meaning of this event is the result of the Cholesterol test is $51\mu /L$, which reflects an abnormal health status.
%$e = (a_1,a_2,...,a_n)$ is an electric record containing $n$ attributes. The attributes can be categorical or numerical and the $i$-th attribute is $A_i$, the value of which is $a_i$.
An episode of a patient EHR data is a clinical event sequence, which may consist of hundreds of clinical events.

\subsection*{Clinical Outcome Prediction}
Clinical outcome prediction is to dynamically predict whether a clinical outcome will happen in 24 hours based on an episode of a patient. 
We aim to dynamically predict two outcomes in this work. 
In the first ``death prediction task'', the  outcome is death in hospital or discharge to home. 
In the second ``ICU admission prediction task'', the  outcome is clinical deterioration (need to be immediately transformed to ICU), or stable clinical  status.

%A heterogeneous event sequence consists of a list of heterogeneous events, each of which must has an common attribute ``record time'' called ``t''. Formally, it is arranged in an ascending sort order by ``$t$'' as $e_1,e_2,...,e_T$ where $e_t = (a_{1}^{i},...,a_{n}^{i},t^i)$.
%We denote the event sequence in a period of time $[T_{start},T_{end}]$ as $\{e_t\}_{T_{start}\leq e_t.t\leq T_{end}}$

\begin{table}[!ht]

    \renewcommand\arraystretch{1}
    \centering
    \newcolumntype{e}{m{6cm}}
    \newcommand{\Multirow}[2]{\multirow{#1}{6cm}{#2}}
    \newcolumntype{a}{m{3cm}}
    \newcolumntype{n}{m{2cm}}
    \newcolumntype{v}{>{\centering\arraybackslash}m{2cm}}
    \newcommand{\twol}[2]{\makecell{#1 \\ #2}}
    \newcommand{\Chline}{\cmidrule{2-5}}
    
    \begin{tabular}{envvv}
        \toprule
            \textbf{event descriptions}    & 
            % \textbf{attribute list} & 
            \textbf{\twol{event name}{(top 2)}} & 
            \textbf{\twol{frequency}{(rank-in-all)}} &
            \textbf{\twol{coverage}{(rank-in-all)}}  & 
            \textbf{\twol{frequency}{per patient}} \\
        % \endfirsthead
        \midrule[\heavyrulewidth]
        \addlinespace
            \Multirow{2}{\textit{Chart events} include routine vital signs, ventilator settings, mental status, and so on.} 
            & Heart Rate & \twol{5171250}{(0.01\%)} & \twol{0.64}{(0.16\%)} & 173.4 \\ \Chline
            & SpO2 & \twol{3410702}{(0.01\%)} & \twol{0.479}{(0.33\%)} & 153.0 \\ % \Chline
            % & Respiratory Rate & \twol{3381160}{(0.02\%)} & \twol{0.479}{(0.33\%)} & 151.7 \\ % \Chline
            % & value(bigeminy, \newline fusion beats, \newline nodal bigeminy ...)  
            % & Ectopy Type & \twol{3242818}{(0.02\%)} & \twol{0.482}{(0.26\%)} & 144.6 \\
    
        \midrule[\heavyrulewidth]
        \addlinespace
            \Multirow{2}{\textit{Input events} are any fluid which have been administered to the patient: such as oral or tube feedings or intravenous solutions containing medications.}
            & 0.9\% Normal Saline & \twol{2363812}{(0.05\%)} & \twol{0.393}{(0.66\%)} & 129.2\\ \Chline
            & Propofol & \twol{369103}{(0.81\%)} & \twol{0.217}{(1.45\%)} & 36.5\\ % \Chline
            % & Dextrose 5\% & \twol{406345}{(0.74\%)} & \twol{0.298}{(1.04\%)} & 29.3\\ % \Chline
            % & duration, amount & Phenylephrine & \twol{93571}{(2.07\%)} & \twol{0.099}{(2.32\%)} & 20.3\\
        %\addlinespace[1cm]

        % \midrule[\heavyrulewidth]
        % \addlinespace
        %     \Multirow{3}{\textit{Datetime events} contain all date measurements about a patient in the ICU. For example, the dialysis.}
        %     & - & Equip Change [MM] & \twol{204762}{(1.24\%)} & \twol{0.037}{(4.02\%)} & 118.9\\ \Chline
        %     & - & INV\#1 Tubing Change [MM] & \twol{190605}{(1.32\%)} & \twol{0.155}{(1.92\%)} & 26.4\\ \Chline
        %     & - & Arterial line Insertion Date & \twol{182011}{(1.35\%)} & \twol{0.188}{(1.68\%)} & 20.8\\
    
        \midrule[\heavyrulewidth]
        \addlinespace
            \Multirow{2}{\textit{Lab events} contain all laboratory measurements for a given patient.} %including out patient data.
            & Hematocrit& \twol{881846}{(0.22\%)} & \twol{0.976}{(0.01\%)} & 19.4\\ \Chline
            & Potassium& \twol{845825}{(0.23\%)} & \twol{0.886}{(0.05\%)} & 20.5\\ % \Chline
            % & SODIUM & \twol{808489}{(0.24\%)} & \twol{0.886}{(0.05\%)} & 19.6\\ %\Chline
            % & value, flag(abnormal, delta, normal) & CREATININE & \twol{797476}{(0.25\%)} & \twol{0.846}{(0.07\%)} & 20.2\\ \Chline
            % & value, flag(abnormal, normal) & CHLORIDE & \twol{795568}{(0.25\%)} & \twol{0.886}{(0.05\%)} & 19.3\\
                
        \midrule[\heavyrulewidth]
        \addlinespace
            \Multirow{2}{\textit{Procedure events} contain procedures for patients.}
            & Chest X-Ray & \twol{32723}{(3.2\%)} & \twol{0.204}{(1.52\%)} & 3.44\\ \Chline
            & EKG & \twol{13962}{(4.35\%)} & \twol{0.167}{(1.82\%)} & 1.79\\ % \Chline
            % & Arterial Line & \twol{12703}{(4.46\%)} & \twol{0.189}{(1.66\%)} & 1.44\\
                
        \midrule[\heavyrulewidth]
        \addlinespace
            \Multirow{2}{\textit{Output events} are fluids which have either been excreted by the patient, such as urine output, or extracted from the patient.}
            & Chest Tubes CTICU CT 1 & \twol{151766}{(1.57\%)} & \twol{0.098}{(2.67\%)} & 33.2\\ \Chline
            & Urine & \twol{107465}{(1.93\%)} & \twol{0.075}{(3.05\%)} & 30.8\\ % \Chline
            % & Chest Tube \#1 & \twol{81128}{(2.23\%)} & \twol{0.071}{(3.1\%)} & 24.5\\
        \bottomrule
        \end{tabular}
\vspace{-8pt}
\caption{Statistics of High Frequency Examples in Different Types of Clinical Events}
\label{tab:clinical events}
\vspace{0.2cm}
\end{table}

\subsection*{Patient Cohort Setup}
We set up two datasets from one real clinical data source, 
MIMIC-III~\autocite{johnson2016mimic} (Medical Information Mart for Intensive Care III), which is a large, freely-available database comprising de-identified health-related data associated with over forty thousand patients who stayed in intensive care units of the Beth Israel Deaconess Medical Center between 2001 and 2012.

We extract 18192 kinds of clinical events with their attributes from the database to get event sequences of patients (the events with top frequency are listed in Table \ref{tab:clinical events}).
%Then the events are assigned according to its belonging to each patient’s admission in the time order, and we get a set of clinical event sequences.
The events whose frequency is less than 2500 are dropped out. 
And we also drop out the admissions, of which the time span from the beginning to the target clinical outcome is less than 36 hours.
%To make the experiment more practical, the events in the period of 24 hours near the target event are hidden. 
Each input of a sample is an episode of a patient clinical event sequence 24 hours before the target outcome.
%1000 events The remaining event sequences with 949 different kinds of events are labeled according to the target endpoint outcome of the task.
The statistics of final clinical  event sequences in the two tasks are summarized in Table \ref{tab:enropy}.
%, where the percentage in the second column is the positive sample rate.

\begin{table}[!ht]
\renewcommand\arraystretch{1}
\centering
\begin{tabular}{|l|c|c|c|}
\hline
\textbf{Dataset}        & \textbf{\# of samples}  & \textbf{\# of events}    & \textbf{Avg timespan}\\
\hline

death      & 24301(8\%) & 20290879& 3d 15h 58m\\
\hline
ICU admission & 19451(21\%) & 14515198& 4d 18h 31m \\

%Retrieval$^\dag$    & 9.427\\% Retrieval is not appropriate for comparison because short, meaningless replies are filtered out in advance
\hline
\end{tabular}
\vspace{-6pt}
\caption{Statistics of  the  datasets (the percentage in the second column is the positive sample rate)}
\label{tab:enropy}
\end{table}

%The decision time sequence is defined as an equal difference progression. 
%The last item is the time 24 hours ahead of the target event, and the difference is 12 hours.

%\subsection*{Clinical endpoint prediction task}

%The clinical endpoint prediction task is formulated as follow: given a clinical heterogeneous event sequence $\{e_t\}_{T_{start}\leq e_t.t\leq T_{end}}$, and a sequence of time stamps $T_{decision}=\{t_i\}$ called decision time sequence in the following, the problem is, for each $t_i \in T_{decision}$, to predict whether the target event will occur afterwards using $\{e_t\}_{T_{start}\leq e_t.t\leq t_i}$.

\section{Methodology}

\begin{figure}[ht!]
\centering
\vspace{-1mm}
\includegraphics[width=.9\textwidth]{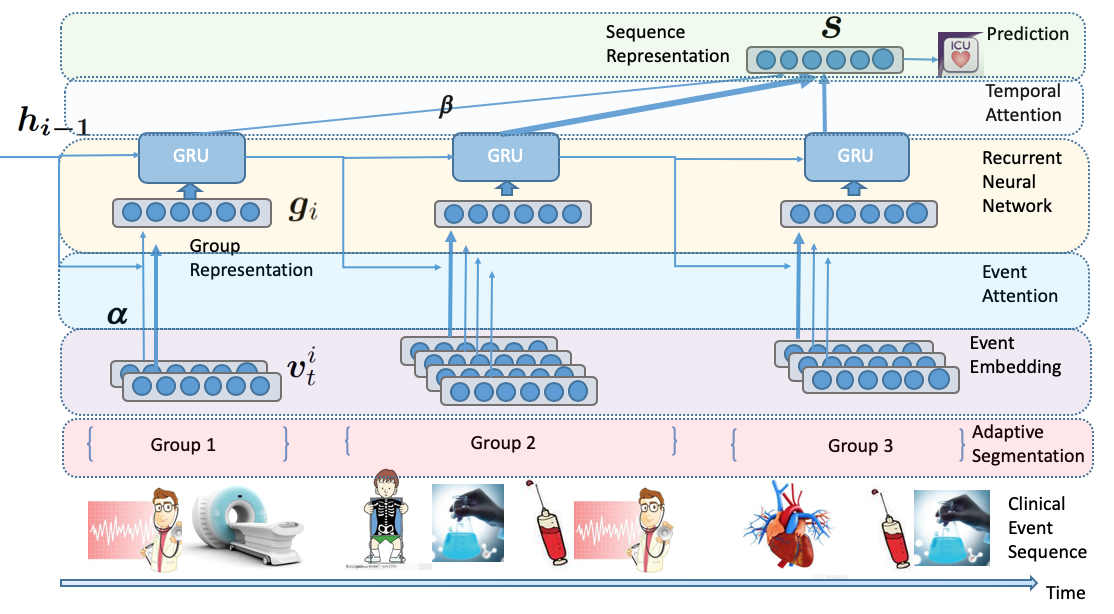}
\vspace{-4mm}
\caption{The overall architecture of our model.  
The original irregular event sequence is segmented into sequential event groups by the adaptive  segmentation module.
Then our model learns the hierarchical representation of the sequential event groups. 
In the low-level representations, each event group is represented as a vector $g_i$ by the event attention. 
In the high-level representations, the embedded sequential groups are modeled by  general recurrent   units (GRU) with inter-group temporal attention. 
}
\label{fig:overview}
\vspace{0.3cm}
\end{figure}

In this section, 
we introduce the technical details of our proposed model. Our model first segments the whole clinical event sequence into several event groups via the adaptive event sequence segmentation module.
Then the model learns hierarchical representations of event sequences
with both event attention and temporal attention mechanisms.
The architecture of our model is illustrated in Figure \ref{fig:overview}.

\subsection{Adaptive  Segmentation}

To distinguish  long-term temporal dependencies from  co-occurrence of important events in short range, we 
adaptively segment an event sequence for a patient into sequential groups, according to the irregular record time of events. 
As events in the same group are exchangeable,   sequential groups can avoid the influence of the  short-range order noisy  in clinical events.
Moreover, sequential groups reduce the length of the sequences fed to RNN, which makes  capturing long-term temporal dependencies easier.

We  find segmentation points of an event sequence by minimizing the maximum time span of the resulting segments.
Formally, given a event sequence $\{e_t\}$, 
$k-1$ segmentation points can split the sequence into $k$ groups $\{G_i\}_{i=1}^k$,
where the event group $G_i =\{e_t | t_{i}'\leq e_t.time\leq t_{i+1}'\} $ is an episode of clinical events from time $t_i'$ to time $t_{i+1}'$, where $e_t.time$ is the record time of the event $e_t$. 
And the time span of a  group is defined as the time difference of the last event and the first in the group, 
namely $span (G_i) = \max_{e_k,e_j\in G_i}\{ e_{k}.time-e_j.time \}$.
So the optimal choice of the segmentation points can be found by minimizing  the following:

$$t_1',...,t_{k+1}'=\arg\min_{t_1',...,t_{k+1}',k \leq M} \max_{1\leq i \leq k}\{span (G_i)\} $$
where $M$ is the max number of groups and $k\leq M$  is the constraint to avoid the segmentation too fine-grained.
%The details of the implementation can be found in supplementary materials.

The adaptive segmentation is designed in a way of the combination of greedy method and binary search. 
We binary search the minimal upper bound of the maximum time span of all groups.
And we then verify the searched upper bound of time spans by trying to greedily construct a solution satisfying
the constraints of $M$ and the time span upper bound. 
The time complexity of the algorithm is $O(L \times \log T)$, where $L$ is the length of the event sequence 
and $T$ is the difference of end time and start time of the sequence. 
%$L$ is set to be a large enough value in our experiences with the two datasets as 10 day, namely 36000 second.
%So $\log T$ will not exceed 16 and
We can regard this algorithm as an algorithm of linear complexity with a big consistent coefficient.

%\subsection{Heterogeneous Event Embedding}
Clinical events have many attributes,  which are not considered in previous popular methods, such as word2vec, GloVe and so on. 
To represent clinical events with their attributes,
we embed each clinical event $e_t$  into the low dimension space as a vector $\bm v_{t}$ in the   way described in the previous work~\autocite{liu2018learning}. The representing vector $v_{t} \in \mathbbm{R}^N$ (where N is the event embedding dimension)
%is mainly decided by event type  and the event attributes, and 
is the sum of event type vector (as basic event information) and event attribute encoding vector (as the description of event feature).

\subsection{Hierarchical Representations with Attention Mechanisms}

Based on the sequential event groups, the model can
learn hierarchical representations to capture long-term temporal dependencies. 
In the low-level model, the model automatically identifies critical clinical events in each group via event attention mechanism and aggregates the events to form event group representations. 
In the high-level model, the meaningful long-term temporal dependencies of clinical  event groups are captured by a recurrent neural network with temporal attention mechanism in the sequence representation.
The hierarchical representations help to learn long-term temporal dependencies in the original event sequences.

\subsubsection{Event Group Representation}
 To select the significant events in each group and compact events in the same group into one vector as the event group representation, the event attention mechanism are added in the low-level model.

Given sequential groups produced by the adaptive segmentation module  
$\{G_i\}_{i=1}^T$, where $G_i =\{ \bm v_{1}^i,...,\bm v_{n_i}^i\}$,
attention score of each event in the group is calculated by the event attention  mechanism.
%is employed to
%merge event embeddings into  group representations. 
The scalars $\alpha_t^i$ are the event attention weights that govern the influence of event embeddings
$\bm v_{1}^i,...,\bm v_{n_i}^i$ in the group $G_i$.

We use a multi-layer perceptron (MLP) with one hidden layer to generate $\alpha_t^i$ based on the event embedding vector $\bm v_{t}^i$ and the hidden state of the previous time as follows:

$$q_{t}^i = \bm w_q \times  \tanh (W_e \times \bm v_{t}^i + W_h \times \bm h_{i-1}+\bm b_h)$$
where $1\leq i \leq T,1\leq t \leq n_i$.
$$\alpha_1^i,...,\alpha_{n_i}^i = \operatorname{softmax }(q^i_{1},..., q^i_{n_t})\space \space \space for\space  1\leq i \leq T$$
where $\bm h_{i-1} \in \mathbbm{R}^S$ is the hidden state of the previous gated recurrent units (GRU) ~\autocite{chung2014empirical}, (which will be described in the following section) $q^i_{t}$ is the latent layer of the event $e_t$ at group $i$  and $W_e\in \mathbbm{R}^{H\times N}, W_h \in \mathbbm{R}^{H\times S},b_h $  and $ w_q \in \mathbbm{R}^H,$ are parameters to learn.
Notice that $H$ is the hidden layer dimension and $S$ is the GRU hidden state dimension.

The resulting attention scores $\bm \alpha$  reflect the importance of each event in a group according to the temporal context of the group. 
Events in the i-th group are weighted averaged with $\bm \alpha^i$ to get the  group representation  $\bm g_i \in \mathbbm{R}^N$ as the input to the $i$-th GRU unit.
$$\bm g_i = \sum^{n_i}_{t = 1} \alpha_t^i \times \bm v^i_{t}$$

\subsubsection{Sequence Representation}
To spot the critical phases over the sequence  for the final decision and capture long-term temporal dependency of event groups, gated recurrent units (GRU) ~\autocite{chung2014empirical} equipped with temporal attention mechanism is employed as the high-level model. 
$$\bm h_i = GRU(\bm{h_{i-1}},\bm g_i,\theta)$$
where the function 
$GRU(\cdot)$ represents the recurrent unit,
which use the previous hidden state $\bm h_{i-1}$ and current input vector $\bm g_i$ to update the hidden state.%which is a shorthand for Eq.(8-12) in the following, 
And $\theta$ represents all the parameters of GRU.

The vector $\bm \beta=(\beta_1,...,\beta_T)$ contains the temporal attention weights of each group in the sequence. 
And we use a fully connected feedforward network to generate $\bm \beta$ from the output of GRU at each time as follow:
$$\bm \beta = \operatorname{softmax} (\bm w_{temporal} \times O_T)$$
where $O_T = (\bm h_1,\bm h_2,...,\bm h_T)\in \mathbbm{R}^{S\times T}$ is the output matrix and $\bm w_{temporal}\in \mathbbm{R}^S$ is a vector of parameters to learn.

The sequence representation $\bm s$  is the weighted average of the output matrix $O_T$. We use $\bm s$ to predict the true label $\hat{y_T}$ of the sequences.
%In this way, important event groups can contribute more to the final prediction.  
$$ \bm s = O_T \times \bm \beta$$
$$y_T = \operatorname{sigmoid} (\bm w_p \times \bm s +b_p )$$
where $\bm w_p \in \mathbbm{R}^S$ and $b_p$ are parameters to learn.

The cross-entropy loss function is used to calculate the classification loss of each sample as follows:
$$\operatorname{Loss}(\{e_t\},\hat{y_T})
=  \hat{y_T}\times \ln y_T  +(1-\hat{y_T})\times \ln(1-y_T)$$
where $\{e_t\}$ is the input event sequence  and $\hat{y_T}$ is the label indicating whether the clinical outcome happens. 
And we can sum up the losses of all the samples in one mini-batch to get the total loss for back propagation.

\section{Results and Discussions}

\subsection{Comparison  Methods and Settings}
% We compare our model with the following models for clinical outcome prediction tasks. 
% These models cover popular models in the literature, including
We compare our model with popular models in the literature, which include
\textit{bag-of-words vector classifiers} (i.e. support vector machine~\autocite{scholkopf2001learning} (SVM), logistic regression~\autocite{yalcin2011gis} (LR),  random forest~\autocite{liaw2002classification} (RF)) and \textit{deep sequential models}, such as RETAIN~\autocite{choi2016retain} and RNN~\autocite{leebig} (implemented with GRU).

Due to the fact that SVM, LR, RF cannot handle a sequence input, the event sequence is compressed into a 0-1 vector in which the i-th element indicates whether the i-th event happens, and then fed into SVM, LR, or RF to make the outcome prediction. We implement these bag-of-words vector classifiers using scikit-learn (https://scikit-learn.org).

Deep sequential models (i.e. RETAIN~\autocite{choi2016retain} and RNN~\autocite{leebig}) take the original event sequence as their inputs as described in section 4.1. We implemented our model and neural network based baselines with Keras (https://keras.io). 
% ablation studies
% For the purpose to evaluate the contributions of the attributes of events, we use two kind of input to feed the proposed model. 

% One(called ``events without attributes'') contains only the information of event types without the information of their attributes, and the other(called ``events with attributes'') contains all information of heterogeneous event sequences.

% We also compare some reduced models of our framework, to check the effect of the attention mechanism. 
% In ``ada-segmentation \& temporal attention'', we remove the event attention from our network. While in ``ada-segmentation \& event attention'', we remove the temporal attention from our network.

%Each dataset is split into 3 parts with fixed proportions, namely training set(60\%), validation set(20\%) and evaluation set(20\%). 
%The data in validation set is used to select hyperparameters of models trained by the training set, in which the samples may be different because of cross-validation. 
%The evaluation set, the details of which is non-transparent for us in the process of training and parameter selection, will then be only used to calculate and report the evaluation metrics for comparison.

% The embedding dimension for heterogeneous events is 32, and the hidden layer of RNN is 64. 
The event embedding size is set to 32 while the hidden layer size is set to 64.
The max number of groups $M$ is set to 32.
When training the models, we used Adam~\autocite{Kingma2014Adam} with the mini-batch of 32 samples and the ``early stopping'' strategy when the performance of validation set drops down.

\subsection{Evaluation Metrics}

%Since all models mentioned above can output predict scores instead of binary labels for 
%the outcome prediction tasks with imbalanced labels
Metrics for binary labels such as accuracy are not suitable for measuring the performance on imbalanced datasets.
Therefore, similar to the works~\autocite{Liu2015Temporal,choi2016retain}, we adopt ROC curves (Receiver Operating Characteristic curves) and PRC (Precision-Recall curves) for evaluation metrics.
Both of these two curves reflect the overall quality of predicted scores, according to their true labels. To get quantitative measurements, the area under ROC(AUC) and the area under PRC(AUPRC) are utilized.

% \begin{figure}[!ht]
% \centering
% \includegraphics[width=.8\linewidth]{resources/result.eps}
% \vspace{-.4cm}
% \caption{AUC and AP of different models on death prediction and ICU-admission}
% \vspace{0.4cm}
% \label{fig:results}
% \end{figure}

\begin{table*}[t]
\renewcommand\arraystretch{1}
\centering
\newcolumntype{P}{>{\centering\arraybackslash}p{1.5cm}}
\begin{tabular}{l@{\hskip 1cm}PP@{\hskip 0.5cm}PP}
\toprule
\multirow{2}{*}{Models} &    \multicolumn{2}{c}{Death} & \multicolumn{2}{c}{ICU admission}  \\
& \textbf{AUC} & \textbf{AUPRC} & \textbf{AUC} & \textbf{AUPRC} \\
\midrule
SVM & 0.7523 & 0.5154 & 0.7973 & 0.7074 \\
LR & 0.8843 & 0.5213 & 0.8734 & 0.7266 \\
Random Forest & 0.8644 & 0.5867 & 0.8389 & 0.8177 \\
\midrule
RETAIN & 0.8967 & 0.5808 & 0.8693 & 0.8029 \\
RNN & 0.9038 & 0.6234 & 0.8636 & 0.8051 \\
Proposed Model &  \textbf{0.9428} & \textbf{0.7476} & \textbf{0.9016} & \textbf{0.8424} \\
\bottomrule
\end{tabular}
\vspace{-5pt}
\caption{performance of different models on death and ICU admission prediction tasks}
\label{tab:results}
\end{table*}

\subsection{Quantitative Results}

Table \ref{tab:results} 
shows the AUC and AUPRC of different models on the death prediction and the ICU admission prediction tasks.  
From the results shown in Table \ref{tab:results}, we can draw the following conclusions:
%our proposed model achieves the best performance on all datasets and all evaluation metrics. 

%To see the effectiveness of heterogeneous event embedding, we can compare the red bars and the yellow bars in Figure.3. 
%Taking the result of death prediction for example, the proposed methods considering the attribute information of events improves the AUC and AUPRC by 3.8\% and 18.0\% compared to the corresponding methods without heterogeneous event embedding module. The improvements of the variations of our model(such as the model without temporal attention module) are 2.7\% and 16.9\%. 
%We can draw the conclusion that by viewing EHR data as heterogeneous event sequences, the performance of the proposed framework and its variations can be improved by the effective usage of the attribute information of events.

% First, sequential models outperform all the non-sequence models.
%and the proposed model obtains the best performance. 
% For example,  on the ``death prediction task'', sequential models(such as RNN, RETAIN and the proposed model) improve the AUC on the death prediction task by at leas 15\%, 3.3\% and 4.6\% compared to SVM, LR and RF. The similar results have been shown in other experiments. We can draw the conclusion that temporal information is effective in endpoint prediction tasks.

First, on the whole, deep sequential models(including RETAIN, RNN and the proposed model) outperform non-sequence models(including SVM, LR, and Random Forest) on both tasks, which suggests that temporal information is effective in the outcome prediction tasks.

Second, our model outperforms all the sequential models. For example, on the ``ICU admission task'', the proposed model improves AUC by at least 3.4\% and AUPRC by at least 3.0\% compared other models on all tasks. 
The improvement verifies our claim that it's more proper to capture temporal dependencies of clinical event sequences in a hierarchical way.
% We can draw the conclusion that hierarchical representations is more effective to model the multi-scale temporal dependency of event sequences of EHR data compared to simple sequential models.

\subsection{Ablation Studies}
In this section, we perform ablation studies to examine the effects of our proposed techniques, namely the event attention mechanism, the temporal attention mechanism, and the adaptive segmentation module.

The first study over two attention mechanisms is performed on both tasks. Specifically, we re-train our model by ablating certain components:

$\bullet$ \textbf{W/O E-Attn}, where no event attention is performed and a group representation is set as the average of the event embeddings in this group.

$\bullet$ \textbf{W/O T-Attn}, where no temporal attention is performed and the sequence representation is set as the final output of the GRU.

\begin{table*}[t]
\renewcommand\arraystretch{1}
\centering
\newcolumntype{P}{>{\centering\arraybackslash}p{1.5cm}}
\begin{tabular}{l@{\hskip 1cm}PP@{\hskip 0.5cm}PP}
\toprule

\multirow{2}{*}{Models} &   \multicolumn{2}{c}{Death} & \multicolumn{2}{c}{ICU admission}  \\
& \textbf{AUC} & \textbf{AUPRC} & \textbf{AUC} & \textbf{AUPRC} \\

\midrule
W/O T-Attn & 0.9348 & 0.7181 & 0.8987 & 0.8400 \\
W/O E-Attn & 0.9170 & 0.6404 & 0.8930 &  0.8376 \\
Full Model & \textbf{0.9428} & \textbf{0.7476} & \textbf{0.9016} & \textbf{0.8424} \\
\bottomrule
\end{tabular}
\vspace{-5pt}
\caption{Ablation study over attention mechanisms}
\label{tab:ablation_attn}
\vspace{-2pt}
\end{table*}

Results of attention mechanism ablation studies are represented in Table \ref{tab:ablation_attn}. We can see that
both attention mechanisms contribute to the strong empirical results of our model represented previously. 
It is noteworthy that the event attention, one of the important part of hierarchical representations, plays a more critical role in our model compared to the temporal attention, especially on the death prediction task. 
% For example, the proposed model improves the AUC and AUPRC of death prediction by 3.1\% and 10.9\% compared to the model without event attention layer, while the improvements are 1.\% and 2.1\% compared to the model without temporal attention.
% We infer the reason is that the event attention makes the model concentrate more on the important events from the long sequence. 
% other than treating all clinical events equally in the groups.

Besides the attention mechanisms, 
study over the adaptive segmentation is performed on the death prediction task. 
We re-train our models by replacing the adaptive segmentation module with the fix-length segmentation which splits the original sequence into groups of equal size events (except the last group). 
Group size of the fix-length segmentation a hyperparameter. Notice that the fix-length segmentation degenerates to no segmentation if group size is set to 1.

\begin{figure}[t]
\centering
\includegraphics[width=.8\linewidth]{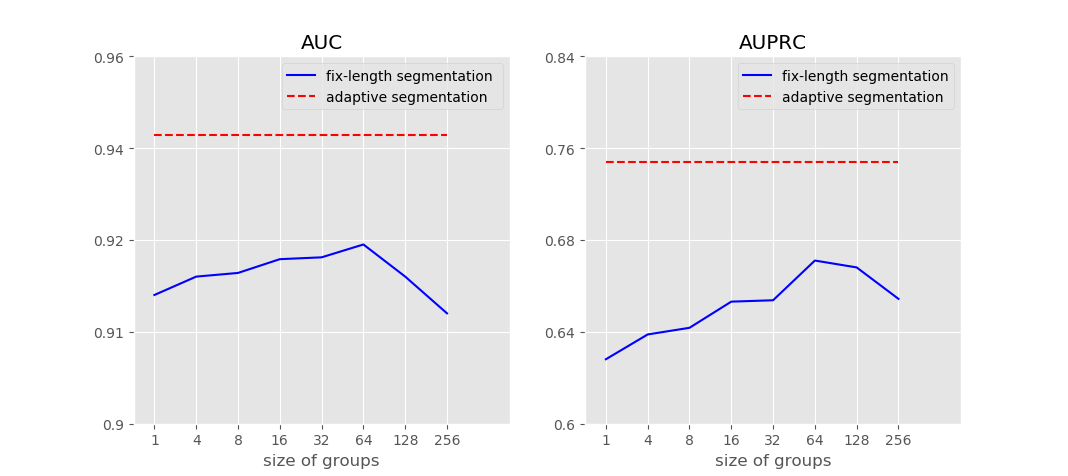}
\vspace{-0.4cm}
\caption{Ablation over the adaptive segmentation on the death prediction task. The fix-length segmentation splits the event sequence into groups of equal size (except the last one).}
\vspace{0.2cm}
\label{fig:varying length}
\end{figure}

Figure \ref{fig:varying length} shows the AUC and AUPRC of the proposed model where the adaptive segmentation is replaced by the fixed-length segmentation with different group sizes in the death prediction task
(the trend in the ICU admission task is similar)
. We can see that the performance goes down when the group size becomes too small or too large. We infer that if the number is too small, local dependencies of events are modeled as long-term dependencies. And if the number is too large, long-term dependency is lost when the corresponding events are assigned to the same group. 
Besides, it's obvious there is a performance gap between the adaptive segmentation and all other segmentation methods, which verifies our claim that the adaptive segmentation can help model long-term dependencies and is suitable for long irregular event sequences.
% the model with the event segmentation outperforms all other segmentation methods by at least  , which indicates the adaptive segmentation is robust and it helps to joint model the local dependency and long-term dependency in the sequence for learning hierarchical representations.

% \subsection{Feature Analysis
\subsection{Important Events}

\begin{table}[t]
\renewcommand\arraystretch{1}
\centering    
\begin{tabular}{p{6cm}>{\centering\arraybackslash}p{2cm}p{4cm}@{\hskip 1cm}>{\centering\arraybackslash}p{2cm}}
    \toprule
        \multicolumn{4}{c}{\textbf{Top Events (Sorted by Median of Event Attention Scores)}} \\ 
    \midrule
        \multicolumn{2}{c}{Death} & \multicolumn{2}{c}{ICU admission} \\
        event & attn score & event & atten score \\
    \cmidrule(lr){1-2} \cmidrule(lr){3-4}
    Blood Products & 0.9965 & Blood PH & 0.9998  \\
    Radiologic Study: thoracic lumbar sac & 0.9896 & Vancomycin & 0.9995\\
    NV\#2 Waveform Appear: overshoot & 0.9713 & Hematocrit (35-51) & 0.9967 \\
    Heart Rhythm & 0.9702 & Edimentation rate & 0.9885\\
    Pain Location: periumbilical & 0.9668 & Daily Weight & 0.9850 \\
    Family Communication & 0.9523 & Bilirubin Total & 0.9834 \\
     
    \bottomrule
\end{tabular}
\vspace{-4pt}
\caption{Top important events on the death prediction task and the ICU-admission prediction task.}
\label{tab:top event}
\end{table}
\vspace{0.3cm}
\vspace{-10pt}

We analyze the events to which our proposed model pays most attention in prediction.
%, based on the event attention scores. 
%The event attention score of an event can measure how much attention the model pays to the event in the prediction.
In particular, we use the median of all event attention scores of an event type on a specific task as the importance of the event type on the task.

Top important events on two tasks are listed in Table \ref{tab:top event}. 
We can see that even though our model mainly focuses on laboratory tests (such as ``Heart Rhythm'' and ``Blood PH'') on both tasks, the specific events attracting the model on two tasks are different due to their different prediction targets.
% We can see that our proposed model is more concerned about laboratory tests(such as ``Heart Rhythm'' and ``Blood PH''). 
It is also perhaps surprising that owing to our data-driven approach, our model can select ``Family Communication'' as an important event type on the death prediction task, which may be ignored by  doctors.
% from a clinical perspective.

% Important events for clinical outcome prediction tasks can be picked out by the event attention mechanism in the proposed model.  
% The same events will get different event attention scores due to the local dependency with their surrounding events and the long-term dependency resulting from their phase in the temporal structure.  
% We divided the samples in each dataset into two groups according to the label to discover some differences of the statistics of the events between patients with different clinical endpoints.
% Thus, we calculated some statistics, including  the average event attention score,  the 1st quartile, the median, the 3rd quartile, the maximum of attention score and so on, of all kinds of events in each group. The top 10 events in each group  with highest median attention score and 3rd Quartile are listed in Table \ref{tab:top event}(for death prediction task) and Table \ref{tab:top event}(for ICU prediction task) respectively.  As the results, the top events in each group are similar both in term of median and the 1st quartile, indicating that these statistics can represent the distribution of attention score of critical events. We can find that important events in different dataset are different, which indicates that our model are able to capture critical information for different end point prediction tasks. 

% \input{discussion}
\section{Conclusion}
In this paper, we proposed a model to learn hierarchical representations of long and irregular clinical event sequences of EHR data for clinical outcome prediction. We validate the performance of our model on real clinical datasets for death and ICU admission prediction tasks. The significant improvements indicated that our model is suitable for irregular timed EHR data and can capture long-term temporal dependencies of clinical event sequences for precise clinical outcome predictions.

\section*{Acknowledgements}
This paper is partially supported by Beijing Municipal Commission of Science and Technology under Grant No. Z181100008918005, National Key Research and Development Program of China with Grant No. SQ2018AAA010010, and the National Natural Science Foundation of China (NSFC Grant No. 61772039 and No. 91646202).

% \printbibliography

% \defbibheading{bibliography}{\centering References}

\begingroup
\setstretch{0.8}
\defbibheading{bibliography}[References]{\section*{\centering #1}}
\printbibliography
\endgroup

% \section{Conclusion}

% Your conclusion goes at the end, followed by References, which must follow the Vancouver Style 
% References begin below with a header that is centered.  
% Only the first word of an article title is capitalized in the References. 

% \makeatletter
% \renewcommand{\@biblabel}[1]{\hfill #1.}
% \makeatother

% \bibliographystyle{unsrt}
% \begin{thebibliography}{1}
% \setlength\itemsep{-0.1em}

% \bibitem{ref1}
% Pryor TA, Gardner RM, Clayton RD, Warner HR. The HELP system. J Med Sys. 1983;7:87-101.
% \bibitem{ref2}
% Gardner RM, Golubjatnikov OK, Laub RM, Jacobson JT, Evans RS. Computer-critiqued blood ordering using the HELP system. Comput Biomed Res 1990;23:514-28.

% \end{thebibliography}

\end{document}